\documentclass[11pt]{article}

\usepackage[preprint]{acl}
\usepackage{times}
\usepackage{latexsym}
\usepackage[T1]{fontenc}
\usepackage[utf8]{inputenc}
\usepackage{microtype}
\usepackage{inconsolata}
\usepackage{graphicx}
\usepackage{booktabs}
\usepackage{multirow}
\usepackage{array}
\usepackage{placeins}

\newcolumntype{L}[1]{>{\raggedright\arraybackslash}p{#1}}

\title{Beyond Free-Form Generation: Measuring Interface Composition in Competitive NLP Systems}

\author{Xinchen Xiao \\
  Xinjiang University \\
  {\ttfamily xiaoxinchenn@outlook.com}}

\begin{document}
\maketitle

\begin{abstract}
Systems with decoder LLMs can combine free generation with scoring, ranking, or selection. We measure these documented interface combinations with five annotation fields linked to source evidence. The corpora comprise 517 solutions from 42 Kaggle competitions involving NLP and 76 participant systems from 26 SemEval, TREC, and WMT task years. Among eligible generation systems with decoder LLMs, 18 of 55 Kaggle systems and 17 of 20 Shared Task systems are hybrid: free generation coexists with a constrained interface. The remaining systems use free generation alone. Giving each competition or task year equal weight yields hybrid shares of 0.327 and 0.876. Within Eedi, 21 eligible systems with decoder LLMs have five exact interface profiles; ten LLM Science Exam systems share one interface class and have four profiles. Across Kaggle, the eligible systems with decoder LLMs span 18 exact interface profiles, revealing variation within broad task and backbone categories. These estimates characterize documented public systems selected using our selection criteria and include interfaces throughout the complete pipeline, including intermediate stages.
The data are available at \url{https://github.com/xiaoxinchenn/nlp-interface-composition}.
\end{abstract}

\section{Introduction}

Systems with decoder LLMs can use a range of output interfaces. The same model family can implement label scoring through prompts, constrained decoding, retrieval, or reranking \citep{schick2021,gao2021,geng2023,zhu2025}. We use \emph{free generation} for free-form text output and distinguish it from action selection under a declared schema. These interfaces expose different output spaces even when the systems share a model family.

Existing surveys of LLMs in NLP and prompting organize the field primarily by task, prompting strategy, or parameter adaptation \citep{liu2023,qin2026}. Research on shared tasks and machine learning competitions emphasizes evaluation, governance, participant behavior, or leaderboard statistics \citep{parraescartin2021,navamunoz2024,nasios2026}. Neither literature estimates interface composition across documented competition systems with decoder LLMs (Appendix Table~\ref{tab:taxonomy_positioning}). The label \emph{solution using an LLM} therefore leaves the output space and surrounding pipeline unspecified.

Classification and ranking systems can use a generative backbone to score labels or rank candidates. Retrieval systems can combine embedding models, decoder LLMs, cross-encoders, and rules, while generation and agent tasks can include verifiers or constrained scoring. We call a system \emph{hybrid} when its documented outputs include both free generation and at least one constrained interface, such as token logits, a classification head, or a ranking score. The definition covers the complete pipeline, including intermediate interfaces: a system can be hybrid even when its decoder generates the final answer.

We contribute two corpora with annotations linked to source evidence and a framework that separates task, backbone, output interface, adaptation method, and system method before aggregation. The corpora contain 517 Kaggle solutions and 76 Shared Task systems, with 179 records independently annotated by both annotators. The analysis estimates the frequency of three interface classes, identifies 18 exact interface profiles in Kaggle, measures variation across tasks, and compares backbone groups within the same competitions. Among eligible systems with decoder LLMs, 125 of 173 Kaggle systems and 32 of 39 Shared Task systems contain a constrained interface. The measurements also reveal variation within task and backbone categories: Eedi's 21 eligible systems with decoder LLMs exhibit five exact interface profiles, including the constrained only and hybrid classes.

\begin{figure*}[t]
  \centering
  \includegraphics[width=\textwidth]{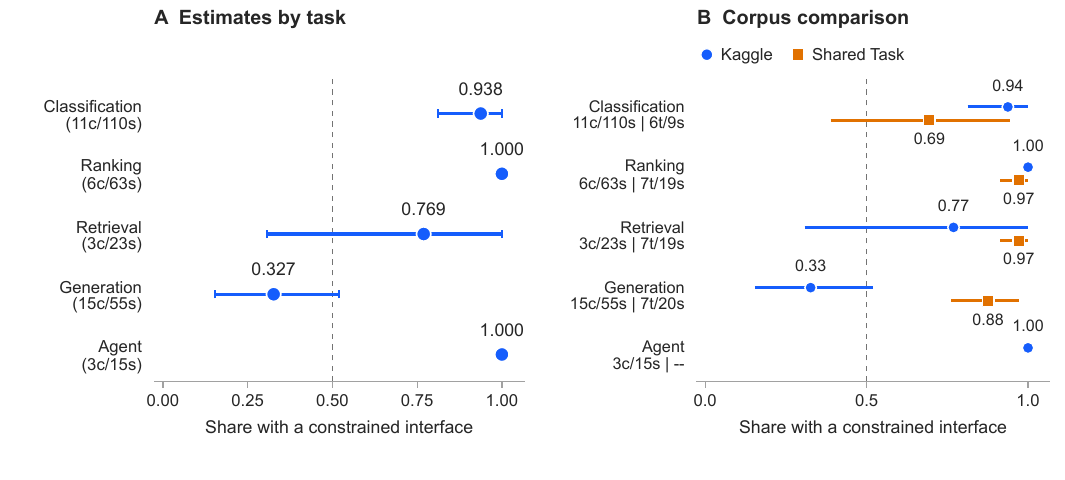}
  \caption{Interface composition across the complete pipeline among eligible documented public systems with decoder LLMs, including intermediate interfaces. (A) Kaggle shares with a constrained interface, giving each competition equal weight, with 95\% intervals from a cluster bootstrap by competition; c/s denotes eligible competitions/solutions. The dashed line marks 0.5. (B) Kaggle (circles) and Shared Task (squares) estimates with separate cluster bootstrap intervals; t/s denotes eligible task years/systems. Shared Task retrieval and ranking use the same records.}
  \label{fig:task}
\end{figure*}

Figure~\ref{fig:task}A reports differences across tasks in Kaggle, and Figure~\ref{fig:task}B compares the corresponding Shared Task estimates. The backbone comparison measures binary interface shares for systems with and without decoder LLMs within competitions that include both groups. Figure~\ref{fig:overview} and RQ3 then relate interface classes to system methods. The corpora retain separate estimands because their sampling units are competitions and task years, respectively.

\section{Related Work}

\paragraph{LLMs in NLP.}
Broad surveys distinguish prompting, use of frozen parameters, and parameter tuning \citep{liu2023,qin2026}, and usually organize evidence around models, methods, or peer-reviewed task studies. Classification through prompting shows that a language model can implement a fixed output interface by scoring label words \citep{schick2021,gao2021}. Research on constrained generation similarly separates the decoder backbone from the permitted output space \citep{willard2023,geng2023,beurerkellner2024,raspanti2025}. Information retrieval research treats LLMs as generators, retrievers, rerankers, and relevance assessors \citep{zhu2025}. These studies describe mechanisms within individual tasks. We estimate how often the corresponding interfaces appear across documented competition systems and compare their frequencies across observed task strata.

Prompt classification, grammar-constrained decoding, embedding retrieval, and generative reranking can all use decoder architectures while exposing different outputs to an evaluator. A free generation interface can also coexist with a verifier, candidate scorer, or rule-based stage in the same pipeline. We therefore code backbone, output interface, and system method independently.

\paragraph{Shared tasks and competitions.}
Research on shared tasks has examined transparency and organizational practice \citep{parraescartin2021}; competition research has studied statistical comparison, ecosystems, incentives, winner characteristics, and the gap between rank and practical utility \citep{navamunoz2024,nasios2026,alroomi2022,ethayarajh2020}. Our unit is the publicly described solution or participant system, and we code five annotation fields from the available evidence. We condition comparisons on task and characterize recurring system designs. The Shared Task corpus applies the same annotation framework to participant papers.

This unit choice separates our study from leaderboard analysis. A leaderboard row is a scored run; a writeup may describe several runs or systems; and one team solution may span several topics. We merge these records at the solution level before coding their components and interfaces. Within Kaggle, rank helps match reports to solutions and cover different leaderboard positions.

\paragraph{Mapping technical practice.}
Systematic mapping and empirical taxonomy development require explicit units, dimensions, stopping rules, and a traceable selection flow \citep{petersen2008,nickerson2013,page2021}. Practitioner reports can support rigorous multivocal synthesis when sources, selection criteria, and quality checks are explicit \citep{garousi2019}. Competition reports describe validation, inference pipelines, and implementation constraints in the grey literature. We therefore link every known label to source evidence and report agreement before adjudication using established reliability measures \citep{cohen1960,artstein2008}. The data card documents sources, curation, intended use, and maintenance \citep{bender2018,gebru2021,pushkarna2022}; GEM and Dynabench illustrate versioned resource reporting \citep{gehrmann2021,kiela2021}.

Prior work supplies methods for documenting datasets, experiments, and competitions, but does not jointly represent task, backbone, output interface, adaptation, and system method at the solution level. Appendix Table~\ref{tab:taxonomy_positioning} compares this coverage directly. We therefore measure these five fields jointly at the solution level.

\section{Research Questions}

We ask: \textbf{RQ1}, what labels appear in the task, backbone, output interface, adaptation method, and system method fields of public competition solutions involving NLP? \textbf{RQ2}, among systems with decoder LLMs, how does documented interface composition differ across task strata and source corpora? This is the primary question. \textbf{RQ3}, how do recurring system methods relate to the observed interface classes? We merge reports that describe the same system, code the five fields independently, and align equal weighting and resampling with the competition or task year sampling unit.

\section{Corpus and Methods}

\subsection{Scope and Solution Resolution}

Kaggle is a competition platform where organizers specify tasks, evaluation metrics, and submission formats. Participants submit predictions or executable solutions for leaderboard evaluation, and public writeups describe their systems. Examples include multiple-choice science question answering and ranking candidate misconceptions in student reasoning.

We screened an official Kaggle directory snapshot of 749 competitions dated August 4, 2026, restricting the frame to competitions launched from 2021 onward. Prespecified NLP task and modality terms, supplemented by searches for generation, retrieval, matching, and agent tasks, produced a registry of 42 competitions. Inclusion required a completed competition, an official task page and leaderboard, language or code as a central modality (or a declared boundary case), and public first-party solution material. Selection preceded annotation and did not use the five fields.

Of the 42 competitions, 20 are core NLP: text is the primary input, and the evaluated target is a standard label, score, span, retrieved or matched item, ranked list, preference, or generated text. Seven are adjacent: the target remains centered on language, but the task includes substantial components involving code, behavior, multimodal matching, query construction, or constraints specific to a domain. Fifteen are boundary cases involving language-based agents, open program or sequence construction, search, or structured reasoning. Multilingual competitions follow the same task and modality criteria. Our target population is the set of publicly documented solutions in this registry.

Each distinct discussion topic was assessed for inclusion based on its text. Inclusion required an eligible report type (solution, implementation, synthesis, structured method, personal result, or official overview), at least two distinct kinds of methodological evidence, and at least 300 substantive characters. Documented exclusion reasons covered questions, placeholders, compilations containing only links, discussions that do not describe solutions, records outside the scope, and insufficient detail.

Applying these rules across the registry produced 517 solution records. We merge posts describing the same solution into one record and split reports describing multiple systems. Leaderboard rank supports report matching and coverage across ranking strata. Each record retains its competition and source IDs, source URL, evidence locators, screening decision, and flags for evidence loss. The data package includes the registry, entity map, a concise annotation guide, evidence locators, final annotations, and aggregate statistics.

\subsection{Sources and Report Quality}

The corpus draws on public Kaggle topics, messages, and writeups. Several topics may describe one solution, and one summary may describe several solutions. We record hashes for source files and text or comment locators, note unavailable posts or referenced materials, and check sources within each sampling stratum.

Known labels require a source locator; absent reporting is coded \texttt{unknown}. Drawing on \citet{garousi2019}, we assess reports by source authority, methodological detail, date and scope, and traceability. The labels describe the system designs documented in the reports.

\subsection{Annotation Framework}

We annotate each solution with five independent, multilabel fields. \emph{Task} covers classification, retrieval, ranking, generation, and related targets; \emph{backbone} identifies encoders, decoder LLMs, embedding models, rerankers, GBDTs, and rules. \emph{Output interface} records heads, token logits, ranking scores, free generation, and action policies. \emph{Adaptation method} covers full fine-tuning, low-rank adaptation (LoRA), quantized LoRA (QLoRA), and no tuning. \emph{System method} records operations across pipeline stages such as retrieval, reranking, ensembling, postprocessing, test-time procedures, verification, and agent planning. Each field also records \texttt{known}, \texttt{unknown}, or \texttt{not\_applicable}; unreported information is coded as unknown. Known labels require source locators, and annotators flag judgments with low confidence separately.

The task and output interface fields were coded independently from separate evidence. The task field records the evaluated target or objective; the output interface field records the interfaces documented in the system report. A solution belongs to the group of systems with decoder LLMs when \texttt{decoder\_llm} appears in its known backbone field. We define three interface classes. The \emph{free generation only} class has an output set of exactly \{\texttt{free\_generation}\}. The \emph{constrained only} class has at least one constrained interface and no free generation, while the \emph{hybrid} class contains both. For the binary analysis, we code the constrained only and hybrid classes as 1 and the free generation only class as 0. This outcome records whether the documented system contains a constrained interface. Because heads, token logits, ranking scores, and action policies differ functionally, we also report the three classes, exact interface profiles, and a sensitivity analysis that treats action policy as an open interface.

Backbone and output interface are multilabel fields coded at the solution level. The output set covers the complete documented pipeline, including intermediate stages. A hybrid system may use a decoder to generate the final answer while another component scores or selects candidates. Identifying the final submission interface or attributing an interface to a decoder requires separate evidence about the relevant stage or component.

The codebook fixes the operational definitions, multilabel rules, evidence requirements, and the distinction among known, unknown, and not applicable.

\subsection{Annotation and Reliability}

A primary annotator coded every record, and a second annotator independently coded a stratified sample of 179 of the 593 systems (30.2\%). The sample covers all 42 Kaggle competitions, all 26 task years in the Shared Task corpus, both source types, and the represented task categories. Both annotators used the same codebook and evidence requirements, and agreement was computed before adjudication. For each dimension, we measured status Cohen's $\kappa$, binary $\kappa$ for each observed label, macro and minimum label $\kappa$, and exact set agreement \citep{cohen1960,artstein2008}. Table~\ref{tab:agreement} reports these measures for the sample annotated by both annotators.

\begin{table}[t]
\centering
\footnotesize
\begin{tabular}{lrrrr}
\toprule
Dimension & Status $\kappa$ & Macro $\kappa$ & Min. $\kappa$ & Exact \\
\midrule
Task & 0.970 & 0.899 & 0.711 & 0.832 \\
Backbone & 0.880 & 0.919 & 0.747 & 0.782 \\
Output & 0.848 & 0.875 & 0.733 & 0.816 \\
Adaptation & 0.951 & 0.963 & 0.876 & 0.944 \\
System & 0.814 & 0.852 & 0.668 & 0.749 \\
\bottomrule
\end{tabular}
\caption{Agreement before adjudication on 179 system records independently annotated by both annotators. Status $\kappa$ is Cohen's $\kappa$ over known/unknown/not applicable; Macro and Min. are the mean and minimum binary $\kappa$ across observed labels; Exact is exact set agreement.}
\label{tab:agreement}
\end{table}

Across the five fields, status $\kappa$ ranges from 0.814 to 0.970 and the mean $\kappa$ across labels ranges from 0.852 to 0.963. Exact set agreement ranges from 0.749 to 0.944. Minimum $\kappa$ values across labels are lower for sparse labels, which the mean across labels can obscure. The derived interface classifications, including eligibility states, agree on 163 of 179 records (91.1\%). Annotators resolved disagreements by consulting the linked evidence, and the analysis uses the consensus labels. Every known consensus label retains a source locator within the study scope; sensitivity analyses exclude or recode backbone and output fields with low confidence.

\subsection{Estimand and Uncertainty}

The RQ2 analysis includes solutions whose known backbone set contains \texttt{decoder\_llm} and whose known output set is nonempty. Within this subset, we report an aggregate estimate that weights competitions equally and contrasts between tasks. We first calculate the share with a constrained interface within each eligible competition and then take the unweighted mean across competitions. Each competition receives equal weight, regardless of how many public reports it contains.

Formally, let $C$ be the eligible competitions, $n_c$ the eligible solutions in competition $c$, and $y_{ci}=1$ when eligible system $i$ contains at least one constrained interface. We estimate
\[
\hat{\theta}=\frac{1}{|C|}\sum_{c\in C}\left(\frac{1}{n_c}\sum_{i=1}^{n_c}y_{ci}\right).
\]
This estimand is the mean, across competitions, of the share of eligible documented public systems selected using our selection criteria whose complete pipeline contains at least one constrained interface. It differs from the pooled share across solutions because each competition receives the same weight. Estimates by task replace $C$ with competitions supporting that task; solutions with multiple task labels can enter multiple task strata and enter the overall estimate once.

Task estimates require at least two competitions and four eligible solutions. The Kaggle analysis applies this fixed support rule and outcome definition to all 517 records.

We estimate uncertainty with 10,000 cluster bootstrap replicates at a fixed seed, resampling competitions and retaining all their eligible solutions. We apply the same design within each supported task. Three descriptive contrasts compare classification, ranking, and retrieval with generation. For each contrast, we resample the union of competitions supporting either task, preserve multilabel membership, and subtract the generation estimate. We also report pooled weighting of solutions, scope breakdowns, sample support by task and scope, and estimates after excluding records with low confidence in the relevant field. These descriptive intervals summarize variation across the selected competitions with the selection criteria held fixed. For strata supported by fewer than five competitions, the intervals primarily reflect variation among the observed competitions.

As a backbone comparator, we apply the same three interface classes to systems whose known backbone set does not contain \texttt{decoder\_llm}. We retain only competitions containing at least one eligible system from each backbone group, which defines the common support subset. Within each competition, we calculate $d_c=p_{c,D}-p_{c,N}$, where $D$ denotes systems with decoder LLMs and $N$ denotes systems without decoder LLMs. We average $d_c$ equally across competitions. A paired cluster bootstrap resamples these competitions while retaining both group shares together. We also omit one competition at a time to assess dependence on individual competitions.

Equal weighting and cluster resampling reduce sensitivity to competitions with many reports. The outcomes describe interface composition; rank supports report matching and sample coverage.

\subsection{Complementary Shared Task Corpus}

Official overview and findings papers defined a candidate frame of 527 participant papers across 26 SemEval, TREC, and WMT task years from 2023--2025. We required a substantive participant report associated with an official task year. Selection covered strata with high, middle, and lower leaderboard ranks within each task year, using a fixed order for replacement reports and leaving a stratum short when no eligible report was available. These strata describe relative leaderboard positions within each task. The resulting corpus contains 76 participant systems.

Official task metadata supplies task labels; participant papers supply the other four fields under the same codebook. Every known label links to evidence located on specific pages. We checked field values and source locators for all 76 records. The Shared Task analysis uses the same decoder eligibility rule and interface outcome and treats each task year as one equally weighted unit and one bootstrap cluster. Supported strata require at least two task years and four eligible systems. Ten thousand cluster bootstrap replicates resampled by task year summarize variation across selected task years. A sensitivity analysis by selection stratum reports estimates for the high, middle, and lower groups and omits each group in turn. We analyze the Kaggle and Shared Task corpora separately.

\section{Results}

\subsection{Corpus Coverage and Field Frequencies (RQ1)}

\begin{figure*}[t]
  \centering
  \includegraphics[width=\textwidth]{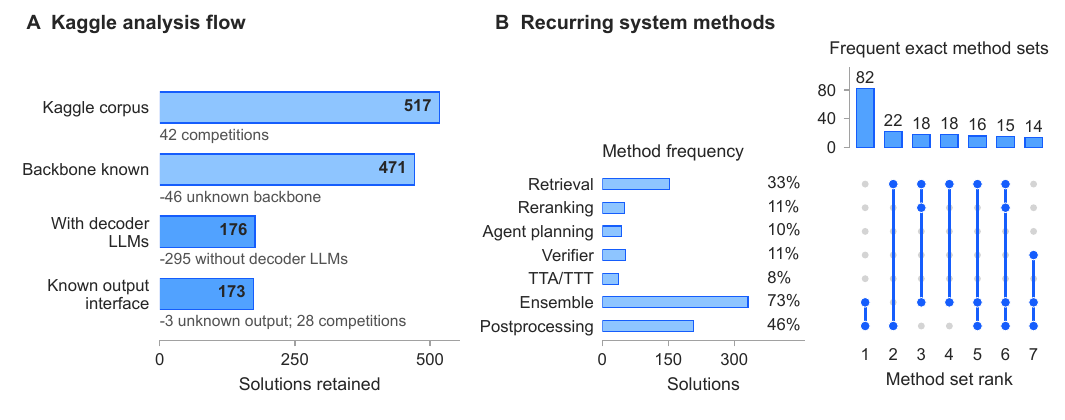}
  \caption{Kaggle analysis flow and recurring system methods. (A) The 517-record corpus yields 176 solutions with decoder LLMs and 173 with known output interfaces. (B) Marginal method frequencies and the seven most frequent exact method sets among 454 solutions with known system methods. Linked filled dots mark methods in each set; bars show solution counts. TTA/TTT denotes test-time augmentation or training.}
  \label{fig:overview}
\end{figure*}

The Kaggle corpus contains 517 solution records across 42 competitions. Of these, 176 have a known backbone set that includes \texttt{decoder\_llm}, and 173 also have a known output interface (Figure~\ref{fig:overview}A). These 173 solutions span 28 competitions. Backbone and system method fields are known for 471 and 454 solutions, respectively; output interface and adaptation fields are unknown for 73 and 234. Appendix Table~\ref{tab:appendix_evidence}B reports the full field status counts.

The five fields capture distinct properties of the documented systems. Classification is assigned to 188 solutions, ranking to 105, retrieval to 80, generation to 103, and agent behavior to 15. Among known backbone fields, encoders appear in 234 solutions and decoder LLMs in 176, while rules (202), embedding models (103), and GBDTs (88) remain substantial components. Among known output interface fields, the most frequent labels are classification heads (146), free generation (121), ranking scores (118), regression heads (72), token classification (69), and token logits (52). Among known adaptation fields, full fine-tuning appears in 113 records, no tuning in 67, LoRA in 43, and QLoRA in 30. Adaptation remains separate because parameter updating is distinct from architecture and output interface; the counts summarize documented methods, with 234 records marked unknown. Because the fields are multilabel, these counts include systems that combine model families and interfaces. The next analysis focuses on interface composition among systems with decoder LLMs.

\subsection{Interface Composition by Task (RQ2)}

\begin{table}[t]
\centering
\footnotesize
\setlength{\tabcolsep}{3.2pt}
\textbf{(A) Generation strata}\par\vspace{0.3em}
\begin{tabular}{lrrrr}
\toprule
Corpus & Systems & C & H & F \\
\midrule
Kaggle & 55 & 0 & 18 & 37 \\
Shared Task & 20 & 0 & 17 & 3 \\
\bottomrule
\end{tabular}
\par\vspace{0.6em}
\textbf{(B) Profiles within competitions}\par\vspace{0.3em}
\begin{tabular}{lrrrrr}
\toprule
Competition & Systems & Profiles & C & H & F \\
\midrule
Eedi & 21 & 5 & 10 & 11 & 0 \\
LLM Science Exam & 10 & 4 & 10 & 0 & 0 \\
\bottomrule
\end{tabular}
\caption{Interface variation among eligible systems with decoder LLMs. C, H, and F denote the constrained only, hybrid, and free generation only classes. (A) Pooled counts for generation systems; equally weighted hybrid shares are 0.327 for Kaggle and 0.876 for the Shared Task corpus. (B) Exact interface profiles within two competitions.}
\label{tab:interface_detail}
\end{table}

We distinguish the constrained only, hybrid, and free generation only classes before combining the first two for the task contrasts below.

The share with a constrained interface differs sharply across the observed task strata (Figure~\ref{fig:task}). The equally weighted competition estimate is 0.938 (95\% interval: 0.813--1.000; 11 competitions/110 systems) for classification, 1.000 (1.000--1.000; 6/63) for ranking, 0.769 (0.308--1.000; 3/23) for retrieval, and 0.327 (0.154--0.519; 15/55) for generation. The differences from generation for classification, ranking, and retrieval are 0.611 (0.375--0.815), 0.673 (0.479--0.845), and 0.443 ($-0.067$--0.797), respectively. The first two intervals remain above zero; retrieval is less precise because it rests on three competitions. All six observed ranking competitions have a share of 1.000. The 1.000--1.000 interval reflects identical shares across the observed competitions. Under the primary coding of declared action policies as constrained, the agent estimate is also 1.000 across three competitions and 15 systems.

The three interface classes comprise 91 constrained only, 34 hybrid, and 48 free generation only Kaggle systems. The most frequent exact interface profiles are free generation (48), classification head (22), classification head plus ranking score (15), classification head plus ranking score and token logits (13), and token logits (12). Action policy plus free generation also appears 12 times. The released annotations record all 18 profiles. Excluding records with a backbone or output field with low confidence leaves every supported task estimate unchanged; classification and ranking retain 103 and 56 solutions, while retrieval, generation, and agent retain all eligible records. With action policies coded as constrained, generation and agent estimates are 0.327 and 1.000; coding them as open yields 0.210 and 0.048, respectively. Classification, ranking, and retrieval remain unchanged. The agent estimate therefore depends strongly on the action policy coding choice.

Every generation system with a constrained interface is hybrid (Table~\ref{tab:interface_detail}A). Kaggle has 18 systems in the hybrid class and 37 in the free generation only class; the Shared Task corpus has 17 and 3, respectively. These are pooled counts. Giving each competition or task year equal weight yields hybrid shares of 0.327 and 0.876.

Task and output interface labels describe different aspects of a system. Classification co-occurs with classification heads in 129 Kaggle records, ranking scores in 84, token logits in 51, and free generation in 25. Ranking and retrieval co-occur with free generation in 12 and 24 records; generation co-occurs with free generation in 92 and action policies in 10. The task field records the evaluated target, while the output interface field records the interfaces documented in the system report. Of 427 solutions with both fields known, 159 have multiple task labels. Both fields allow multiple labels, so these counts describe overlapping groups of solutions.

Registry scope also varies by task. Fourteen of the 15 competitions supporting generation are boundary cases, while classification and ranking are concentrated in core NLP; retrieval has three core NLP competitions. These counts define the support behind the estimates by task. Within the observed registry, classification and ranking systems mostly fall in the constrained only or hybrid classes, whereas generation systems more often use free generation alone.

Within Eedi, 21 eligible systems with decoder LLMs have five exact interface profiles: ten systems are in the constrained only class and eleven are hybrid (Table~\ref{tab:interface_detail}B). LLM Science Exam's ten eligible systems with decoder LLMs are all in the constrained only class and have four exact interface profiles. Exact interface profiles therefore reveal variation under shared competition requirements, within one backbone category and even one interface class.

\subsection{Comparison across Corpora}

The Shared Task corpus contains 42 systems with decoder LLMs, 39 with known output interfaces. Of these, three are in the constrained only class, 29 are hybrid, and seven are in the free generation only class. The equally weighted estimate across task years is 0.805 (95\% interval: 0.656--0.937). Omitting one task year at a time yields estimates between 0.794 and 0.850. Three missingness and confidence analyses yield 0.748--0.779. Estimates for the high, middle, and lower selection strata are 0.786, 0.885, and 0.778; omitting any one stratum yields 0.750--0.869. The aggregate pattern persists when each selection stratum is omitted in turn.

Estimates by task differ across the two corpora. Shared Task estimates are 0.694 for classification, 0.971 for retrieval and ranking, and 0.876 for generation, so the generation share exceeds the classification share. Generation illustrates the difference most clearly: systems in this task category mainly use free generation alone in Kaggle but are frequently hybrid in the Shared Task corpus. The most frequent exact interface profile in the Shared Task corpus is \texttt{free\_generation} plus \texttt{ranking\_score} (21 of 39); 23 systems contain both interfaces. Even task years labeled only as generation have an equally weighted estimate of 0.833, and task years labeled as retrieval, ranking, and generation reach 0.933. The same broad task label can therefore accompany different documented interface profiles.

\subsection{Backbone Comparator and Aggregate Pattern}

Thirteen Kaggle competitions contain eligible systems from both backbone groups. Within this common support subset, the equally weighted shares with a constrained interface are 0.747 for systems with decoder LLMs (69 solutions) and 0.769 for systems without decoder LLMs (111 solutions). The paired difference ($D$ minus $N$) is $-0.022$ (95\% interval from a paired cluster bootstrap: $-0.159$--0.092). Omitting one competition at a time gives differences from $-0.057$ to 0.033. This difference close to zero shows little separation by backbone group in the observed binary interface share. Systems with decoder LLMs also span 18 exact interface profiles, so output interface annotation captures variation that the backbone label omits. Appendix Table~\ref{tab:appendix_evidence}C reports every paired competition.

Across all eligible Kaggle competitions, the equally weighted estimate is 0.615 (95\% interval: 0.454--0.768), and the pooled share across solutions is 0.723. Contrasts between tasks, the backbone comparator, and Shared Task estimates reveal variation hidden by either aggregate.

\subsection{Recurring System Methods (RQ3)}

System methods are known for 454 Kaggle solutions. Ensembling appears in 331 (72.9\%) across 37 competitions; postprocessing in 208 (45.8\%) across 34; and retrieval in 152 (33.5\%) across 20 (Figure~\ref{fig:overview}B). Reranking appears in 50 solutions across seven competitions. By task, ensembling appears in 140 of 183 classification systems and 41 of 87 generation systems; retrieval appears in 92 of 183 and 22 of 87, respectively. These counts include all backbone families with known task and system method fields; task groups may overlap. The most common exact method sets are ensemble with postprocessing (82 records), postprocessing with retrieval (22), ensemble with reranking and retrieval (18), and ensemble with retrieval (18).

Method frequencies differ across the three interface classes. Among eligible systems whose backbones include decoder LLMs and whose system methods are known, retrieval appears in 16 of 33 hybrid systems and reranking in 12. Among 45 systems in the free generation only class, retrieval appears in 15 and reranking in none. Systems in the constrained only class most often use ensembling (71 of 88), whereas systems in the free generation only class most often use postprocessing (32 of 45). Systems in the free generation only class can also contain retrieval, verification, and postprocessing. Interface class therefore adds information beyond pipeline depth.

\section{Discussion}

The Kaggle results show how interface composition varies across tasks among systems with decoder LLMs. Classification and ranking systems commonly document scoring interfaces, whereas generation systems more often use free generation alone. The contrasts between classification and generation and between ranking and generation remain large across the observed registry. Task structure, required submission format, and evaluation metric vary together: label or score submissions partly explain the high shares with a constrained interface in classification and ranking. Exact interface profiles within a competition show how systems vary under shared requirements.

Shared Task systems show a contrasting generation pattern. The generation estimate is 0.876, and the most frequent exact interface profile combines \texttt{free\_generation} with \texttt{ranking\_score}. Across both corpora, 63 hybrid systems document free generation alongside another output interface. Systems across the 18 exact interface profiles in Kaggle also vary in retrieval, reranking, ensembling, verification, and postprocessing, while methods with multiple stages occur in free generation only systems as well. Recording interface composition across the complete system adds detail beyond task labels, decoder backbones, and pipeline depth.

The findings motivate a concrete reporting standard. Public system descriptions should separate model checkpoints, parameter updates, intermediate generation or retrieval, the final output interface, and postprocessing. They should describe model and data sources, validation and search methods, and the components used in the submitted system. Transparency checklists for shared tasks and data cards provide practical precedents \citep{parraescartin2021,pushkarna2022}.

\section{Conclusion}

Among documented public systems selected using our selection criteria, task and backbone labels leave substantial variation in interface composition across the complete system. In the generation strata, 18 of 55 Kaggle systems and 17 of 20 Shared Task systems are hybrid; the rest use free generation alone. The corresponding hybrid shares are 0.327 and 0.876 when competitions or task years receive equal weight. Within Eedi, 21 eligible systems with decoder LLMs have five exact interface profiles, spanning the constrained only and hybrid classes. In LLM Science Exam, ten systems in the constrained only class have four exact interface profiles. Across Kaggle, the 18 exact interface profiles and three interface classes distinguish compositions within broad task and backbone categories. These estimates capture interfaces throughout the documented pipeline, including intermediate stages, under the submission requirements and reporting practices of each corpus.

\section*{Limitations}

The estimates characterize documented public systems selected using our criteria for competitions and reports. Public writeups may overrepresent stronger or more complex systems. This selection issue also arises in other nonprobability studies of NLP communities \citep{michael2023}. We report separate estimates for Kaggle and Shared Tasks, using competitions and task years as the respective sampling units.

Labels describe the system designs documented in public reports. The output interface field is unknown for 73 Kaggle solutions, and the adaptation field is unknown for 234; three solutions with decoder LLMs lack a known output and are excluded from RQ2.

Task and registry scope are correlated in the Kaggle corpus, which includes one core NLP generation competition. Participant papers and Kaggle reports also differ in format and in how much they document intermediate ranking, retrieval, or verification. Accordingly, the comparison across corpora characterizes documented interface composition across these reporting settings. The Shared Task analysis operates at task year granularity, so RAG systems spanning multiple tracks can carry several task labels. Backbone and output interface are recorded for each solution; the analysis measures interface composition across the complete system. Kaggle task contrasts and Shared Task estimates provide complementary views of interface composition.

\section*{Ethical Considerations}

The source material consists of public competition reports and participant papers. The release follows the licensing and redistribution terms of the source material, consistent with lifecycle analyses of dataset context and reuse \citep{paullada2021}. The accompanying data package contains final annotations produced by this project, project and topic identifiers without personal data, source URLs, a concise annotation guide, aggregate statistics, and checksums. It excludes raw API responses, extracted text, comments, images, author names, team names, and other identity fields. The analysis focuses on system designs, and annotators are identified by anonymous roles.

The principal release risk is incorrect attribution or decontextualization. We reduce this risk by linking annotations to source evidence, using neutral language, releasing derived data, and providing a correction or takedown process. Derived annotations and metadata support the reported numerical analyses. The license documentation distinguishes data and documentation produced by this project from third-party sources and records each component's license status. Third-party source material retains its original terms.

\section*{Acknowledgments}

The author thanks the anonymous reviewers and the area chair for their constructive feedback. The author is especially grateful to a friend who wishes to remain anonymous for independently annotating a subset of the systems in this study.

\bibliography{references}

@article{alroomi2022,
  title = {Fathoming Empirical Forecasting Competitions' Winners},
  author = {Alroomi, Azzam and Karamatzanis, Georgios and Nikolopoulos, Konstantinos and Tilba, Anna and Xiao, Shujun},
  journal = {International Journal of Forecasting},
  volume = {38},
  number = {4},
  pages = {1519--1525},
  year = {2022},
  doi = {10.1016/j.ijforecast.2022.03.010}
}

@article{artstein2008,
  title = {Inter-Coder Agreement for Computational Linguistics},
  author = {Artstein, Ron and Poesio, Massimo},
  journal = {Computational Linguistics},
  volume = {34},
  number = {4},
  pages = {555--596},
  year = {2008},
  doi = {10.1162/coli.07-034-R2}
}

@article{bender2018,
  title = {Data Statements for Natural Language Processing: Toward Mitigating System Bias and Enabling Better Science},
  author = {Bender, Emily M. and Friedman, Batya},
  journal = {Transactions of the Association for Computational Linguistics},
  volume = {6},
  pages = {587--604},
  year = {2018},
  doi = {10.1162/tacl_a_00041}
}

@inproceedings{beurerkellner2024,
  title = {Guiding {LLM}s the Right Way: Fast, Non-Invasive Constrained Generation},
  author = {Beurer-Kellner, Luca and Fischer, Marc and Vechev, Martin},
  booktitle = {Proceedings of the 41st International Conference on Machine Learning},
  series = {Proceedings of Machine Learning Research},
  volume = {235},
  pages = {3658--3673},
  year = {2024},
  publisher = {PMLR},
  url = {https://proceedings.mlr.press/v235/beurer-kellner24a.html}
}

@article{cohen1960,
  title = {A Coefficient of Agreement for Nominal Scales},
  author = {Cohen, Jacob},
  journal = {Educational and Psychological Measurement},
  volume = {20},
  number = {1},
  pages = {37--46},
  year = {1960},
  doi = {10.1177/001316446002000104}
}

@inproceedings{gao2021,
  title = {Making Pre-trained Language Models Better Few-shot Learners},
  author = {Gao, Tianyu and Fisch, Adam and Chen, Danqi},
  booktitle = {Proceedings of the 59th Annual Meeting of the Association for Computational Linguistics and the 11th International Joint Conference on Natural Language Processing (Volume 1: Long Papers)},
  pages = {3816--3830},
  year = {2021},
  publisher = {Association for Computational Linguistics},
  doi = {10.18653/v1/2021.acl-long.295}
}

@article{garousi2019,
  title = {Guidelines for Including Grey Literature and Conducting Multivocal Literature Reviews in Software Engineering},
  author = {Garousi, Vahid and Felderer, Michael and M{\"a}ntyl{\"a}, Mika V.},
  journal = {Information and Software Technology},
  volume = {106},
  pages = {101--121},
  year = {2019},
  doi = {10.1016/j.infsof.2018.09.006}
}

@inproceedings{geng2023,
  title = {Grammar-Constrained Decoding for Structured {NLP} Tasks without Finetuning},
  author = {Geng, Saibo and Josifoski, Martin and Peyrard, Maxime and West, Robert},
  booktitle = {Proceedings of the 2023 Conference on Empirical Methods in Natural Language Processing},
  pages = {10932--10952},
  year = {2023},
  publisher = {Association for Computational Linguistics},
  doi = {10.18653/v1/2023.emnlp-main.674}
}

@article{gebru2021,
  title = {Datasheets for Datasets},
  author = {Gebru, Timnit and Morgenstern, Jamie and Vecchione, Briana and Vaughan, Jennifer Wortman and Wallach, Hanna and Daum{\'e} III, Hal and Crawford, Kate},
  journal = {Communications of the ACM},
  volume = {64},
  number = {12},
  pages = {86--92},
  year = {2021},
  doi = {10.1145/3458723}
}

@inproceedings{gehrmann2021,
  title = {The {GEM} Benchmark: Natural Language Generation, its Evaluation and Metrics},
  author = {Gehrmann, Sebastian and others},
  booktitle = {Proceedings of the 1st Workshop on Natural Language Generation, Evaluation, and Metrics},
  pages = {96--120},
  year = {2021},
  publisher = {Association for Computational Linguistics},
  doi = {10.18653/v1/2021.gem-1.10}
}

@article{liu2023,
  title = {Pre-train, Prompt, and Predict: A Systematic Survey of Prompting Methods in Natural Language Processing},
  author = {Liu, Pengfei and Yuan, Weizhe and Fu, Jinlan and Jiang, Zhengbao and Hayashi, Hiroaki and Neubig, Graham},
  journal = {ACM Computing Surveys},
  volume = {55},
  number = {9},
  pages = {1--35},
  year = {2023},
  doi = {10.1145/3560815}
}

@inproceedings{kiela2021,
  title = {Dynabench: Rethinking Benchmarking in {NLP}},
  author = {Kiela, Douwe and Bartolo, Max and Nie, Yixin and Kaushik, Divyansh and Geiger, Atticus and Wu, Zhengxuan and Vidgen, Bertie and Prasad, Grusha and Singh, Amanpreet and others},
  booktitle = {Proceedings of the 2021 Conference of the North American Chapter of the Association for Computational Linguistics: Human Language Technologies},
  pages = {4110--4124},
  year = {2021},
  publisher = {Association for Computational Linguistics},
  doi = {10.18653/v1/2021.naacl-main.324}
}

@inproceedings{ethayarajh2020,
  title = {Utility Is in the Eye of the User: A Critique of {NLP} Leaderboards},
  author = {Ethayarajh, Kawin and Jurafsky, Dan},
  booktitle = {Proceedings of the 2020 Conference on Empirical Methods in Natural Language Processing},
  pages = {4846--4853},
  year = {2020},
  publisher = {Association for Computational Linguistics},
  doi = {10.18653/v1/2020.emnlp-main.393}
}

@article{nasios2026,
  title = {The Ecosystem of Machine Learning Competitions: Platforms, Participants, and Their Impact on {AI} Development},
  author = {Nasios, Ioannis},
  journal = {arXiv preprint arXiv:2604.08001},
  year = {2026},
  url = {https://arxiv.org/abs/2604.08001}
}

@article{navamunoz2024,
  title = {Analysis of Systems' Performance in Natural Language Processing Competitions},
  author = {Nava-Mu{\~n}oz, Sergio and Graff, Mario and Escalante, Hugo Jair},
  journal = {Pattern Recognition Letters},
  volume = {186},
  pages = {346--353},
  year = {2024},
  doi = {10.1016/j.patrec.2024.03.010}
}

@inproceedings{michael2023,
  title = {What Do {NLP} Researchers Believe? Results of the {NLP} Community Metasurvey},
  author = {Michael, Julian and Holtzman, Ari and Parrish, Alicia and Mueller, Aaron and Wang, Alex and Chen, Angelica and Madaan, Divyam and Nangia, Nikita and Pang, Richard Yuanzhe and Phang, Jason and Bowman, Samuel R.},
  booktitle = {Proceedings of the 61st Annual Meeting of the Association for Computational Linguistics (Volume 1: Long Papers)},
  pages = {16334--16368},
  year = {2023},
  publisher = {Association for Computational Linguistics},
  doi = {10.18653/v1/2023.acl-long.903}
}

@article{nickerson2013,
  title = {A Method for Taxonomy Development and Its Application in Information Systems},
  author = {Nickerson, Robert C. and Varshney, Upkar and Muntermann, Jan},
  journal = {European Journal of Information Systems},
  volume = {22},
  number = {3},
  pages = {336--359},
  year = {2013},
  doi = {10.1057/ejis.2012.26}
}

@article{parraescartin2021,
  title = {Towards Transparency in {NLP} Shared Tasks},
  author = {Parra Escart{\'i}n, Carla and Lynn, Teresa and Moorkens, Joss and Dunne, Jane},
  journal = {arXiv preprint arXiv:2105.05020},
  year = {2021},
  url = {https://arxiv.org/abs/2105.05020}
}

@article{paullada2021,
  title = {Data and Its (Dis)contents: A Survey of Dataset Development and Use in Machine Learning Research},
  author = {Paullada, Amandalynne and Raji, Inioluwa Deborah and Bender, Emily M. and Denton, Emily and Hanna, Alex},
  journal = {Patterns},
  volume = {2},
  number = {11},
  pages = {100336},
  year = {2021},
  doi = {10.1016/j.patter.2021.100336}
}

@article{page2021,
  title = {The {PRISMA} 2020 Statement: An Updated Guideline for Reporting Systematic Reviews},
  author = {Page, Matthew J. and McKenzie, Joanne E. and Bossuyt, Patrick M. and Boutron, Isabelle and Hoffmann, Tammy C. and Mulrow, Cynthia D. and Shamseer, Larissa and Tetzlaff, Jennifer M. and Akl, Elie A. and Brennan, Sue E. and Chou, Roger and Glanville, Julie and Grimshaw, Jeremy M. and Hr{\'o}bjartsson, Asbj{\o}rn and Lalu, Manoj M. and Li, Tianjing and Loder, Elizabeth W. and Mayo-Wilson, Evan and McDonald, Steve and McGuinness, Luke A. and Stewart, Lesley A. and Thomas, James and Tricco, Andrea C. and Welch, Vivian A. and Whiting, Penny and Moher, David},
  journal = {BMJ},
  volume = {372},
  pages = {n71},
  year = {2021},
  doi = {10.1136/bmj.n71}
}

@inproceedings{petersen2008,
  title = {Systematic Mapping Studies in Software Engineering},
  author = {Petersen, Kai and Feldt, Robert and Mujtaba, Shahid and Mattsson, Michael},
  booktitle = {12th International Conference on Evaluation and Assessment in Software Engineering},
  year = {2008},
  publisher = {BCS Learning \& Development},
  doi = {10.14236/ewic/EASE2008.8}
}

@article{qin2026,
  title = {Large Language Models Meet {NLP}: A Survey},
  author = {Qin, Libo and Chen, Qiguang and Feng, Xiachong and Wu, Yang and Zhang, Yongheng and Li, Yinghui and Li, Min and Che, Wanxiang and Yu, Philip S.},
  journal = {Frontiers of Computer Science},
  volume = {20},
  number = {11},
  pages = {2011361},
  year = {2026},
  doi = {10.1007/s11704-025-50472-3}
}

@inproceedings{pushkarna2022,
  title = {Data Cards: Purposeful and Transparent Dataset Documentation for Responsible {AI}},
  author = {Pushkarna, Mahima and Zaldivar, Andrew and Kjartansson, Oddur},
  booktitle = {Proceedings of the 2022 ACM Conference on Fairness, Accountability, and Transparency},
  pages = {1776--1826},
  year = {2022},
  publisher = {Association for Computing Machinery},
  doi = {10.1145/3531146.3533231}
}

@inproceedings{raspanti2025,
  title = {Grammar-Constrained Decoding Makes Large Language Models Better Logical Parsers},
  author = {Raspanti, Federico and Ozcelebi, Tanir and Holenderski, Mike},
  booktitle = {Proceedings of the 63rd Annual Meeting of the Association for Computational Linguistics (Volume 6: Industry Track)},
  pages = {485--499},
  year = {2025},
  publisher = {Association for Computational Linguistics},
  doi = {10.18653/v1/2025.acl-industry.34}
}

@inproceedings{schick2021,
  title = {Exploiting Cloze-Questions for Few-Shot Text Classification and Natural Language Inference},
  author = {Schick, Timo and Sch{\"u}tze, Hinrich},
  booktitle = {Proceedings of the 16th Conference of the European Chapter of the Association for Computational Linguistics: Main Volume},
  pages = {255--269},
  year = {2021},
  publisher = {Association for Computational Linguistics},
  doi = {10.18653/v1/2021.eacl-main.20}
}

@article{willard2023,
  title = {Efficient Guided Generation for Large Language Models},
  author = {Willard, Brandon T. and Louf, R{\'e}mi},
  journal = {arXiv preprint arXiv:2307.09702},
  year = {2023},
  url = {https://arxiv.org/abs/2307.09702}
}

@article{zhu2025,
  title = {Large Language Models for Information Retrieval: A Survey},
  author = {Zhu, Yutao and Yuan, Huaying and Wang, Shuting and Liu, Jiongnan and Liu, Wenhan and Deng, Chenlong and Chen, Haonan and Liu, Zheng and Dou, Zhicheng and Wen, Ji-Rong},
  journal = {ACM Transactions on Information Systems},
  volume = {44},
  number = {1},
  pages = {1--54},
  year = {2025},
  doi = {10.1145/3748304}
}

\appendix
\raggedbottom
\makeatletter
\setlength{\@dblfptop}{0pt}
\setlength{\@dblfpsep}{8pt}
\setlength{\@dblfpbot}{0pt plus 1fil}
\makeatother

\section{Annotation Framework Detail}

\begin{table}[t]
\centering
\footnotesize
\setlength{\tabcolsep}{3pt}
\begin{tabular}{L{2.15cm}L{4.75cm}}
\toprule
Research family & Primary emphasis and distinction \\
\midrule
LLM method surveys & Organize published studies by task, prompting, and adaptation \citep{liu2023,qin2026}; we separate model family from documented interfaces and system operations. \\
\addlinespace
Methods with constrained interfaces & Study label scoring or decoding restrictions \citep{schick2021,gao2021,willard2023,geng2023}; we estimate prevalence across competitions and exact interface profiles. \\
\addlinespace
Competition studies & Analyze runs, teams, scores, incentives, or transparency \citep{parraescartin2021,navamunoz2024,nasios2026,alroomi2022}; we code five fields linked to source evidence for each solution record. \\
\addlinespace
This study & Links five fields to estimands by competition or task year, estimates across corpora, and a comparison within the common support subset. \\
\bottomrule
\end{tabular}
\caption{Positioning relative to the closest research families. Entries summarize primary analytical emphasis.}
\label{tab:taxonomy_positioning}
\end{table}

\begin{table*}[t]
\centering
\footnotesize
\textbf{(A) Worked annotation examples}\par\vspace{0.4em}
\setlength{\tabcolsep}{4pt}
\renewcommand{\arraystretch}{1.0}
\begin{tabular}{L{1.3cm}L{8.2cm}L{5.15cm}}
\toprule
Example & Coded fields & Why it matters \\
\midrule
Jigsaw-1 & \textbf{Task:} classification; \textbf{backbone:} decoder LLM; \textbf{output:} token logits; \textbf{system:} ensemble & A solution with a decoder LLM documents a fixed classification interface. \\
\addlinespace
MDC-1 & \textbf{Task:} classification, span; \textbf{backbone:} decoder LLM, embedding model, rules; \textbf{output:} free generation; \textbf{system:} retrieval, postprocessing & A solution labeled as classification can record free generation as its only output interface. \\
\addlinespace
Eedi-1 & \textbf{Task:} classification, ranking; \textbf{backbone:} decoder LLM, embedding model, reranker; \textbf{output:} token logits, ranking score, free generation; \textbf{system:} retrieval, reranking, ensemble, postprocessing & Generated and constrained interfaces coexist in one hybrid solution. \\
\addlinespace
Code-2 & \textbf{Task:} generation, agent; \textbf{backbone:} decoder LLM, rules; \textbf{output:} free generation, action policy; \textbf{system:} verifier, agent planning & Free generation and action policy interfaces coexist with verification and planning methods. \\
\bottomrule
\end{tabular}
\vspace{0.8em}
\begin{minipage}[t]{0.39\textwidth}
\centering
\small
\textbf{(B) Field status}\par\vspace{0.15em}
\scriptsize
\setlength{\tabcolsep}{3pt}
\begin{tabular}{lrrrr}
\toprule
Field & Known & Unknown & N/A & Low conf. \\
\midrule
Task & 445 & 71 & 1 & 59 \\
Backbone & 471 & 46 & 0 & 25 \\
Output & 428 & 73 & 16 & 76 \\
Adaptation & 225 & 234 & 58 & 204 \\
System & 454 & 42 & 21 & 25 \\
\bottomrule
\end{tabular}
\end{minipage}
\hfill
\begin{minipage}[t]{0.57\textwidth}
\centering
\small
\textbf{(C) Backbone comparison within competitions}\par\vspace{0.15em}
\scriptsize
\setlength{\tabcolsep}{3pt}
\begin{tabular}{L{3.05cm}rrr}
\toprule
Competition & $D$: $y/n$ & $N$: $y/n$ & $d_c$ \\
\midrule
Show Us the Data & 1/1 & 3/3 & 0.000 \\
CommonLit Readability & 1/1 & 7/7 & 0.000 \\
Deep Past MT & 0/2 & 0/6 & 0.000 \\
Drawing with LLMs & 1/1 & 5/5 & 0.000 \\
Eedi Misconceptions & 21/21 & 2/2 & 0.000 \\
Google Code Golf & 2/5 & 0/1 & 0.400 \\
Kaggle LLM Science Exam & 10/10 & 17/17 & 0.000 \\
Automated Essay Scoring & 1/1 & 11/11 & 0.000 \\
LLM-Generated Text Detection & 6/6 & 20/20 & 0.000 \\
Make Data Count & 5/16 & 3/3 & $-0.688$ \\
PII Detection & 1/1 & 27/27 & 0.000 \\
Santa 2024 & 0/2 & 0/5 & 0.000 \\
Stable Diffusion Prompts & 2/2 & 4/4 & 0.000 \\
\midrule
Equally weighted share / difference & 0.747 & 0.769 & $-0.022$ \\
\bottomrule
\end{tabular}
\end{minipage}
\caption{Appendix evidence. (A) Representative Kaggle annotations showing independent task, backbone, output interface, and system method coding; aliases map to released solution IDs, and adaptation method is omitted for space. (B) Annotation status counts and flags for low confidence for 517 solutions; status columns are mutually exclusive, Low conf. overlaps them, and N/A denotes not applicable. (C) The backbone comparator within the common support subset. D and N denote systems with and without decoder LLMs, respectively; y/n is the count of systems with a constrained interface over eligible systems, and the paired difference is the D share minus the N share.}
\label{tab:appendix_evidence}
\end{table*}

\paragraph{Task.} The evaluated target or objective: classification, regression, token/span prediction, retrieval, ranking, generation, or agent behavior.

\paragraph{Backbone.} Functional model families used by the system: encoder, decoder LLM, embedding model, reranker, GBDT, or rules.

\paragraph{Output interface.} Final or intermediate outputs documented in the system report: classification or regression head, token logits, token classification, ranking score, free generation, or action policy.

\paragraph{Adaptation method.} How model parameters are used or updated: full fine-tuning, LoRA, QLoRA, or no tuning.

\paragraph{System method.} Operations across pipeline stages: retrieval, reranking, ensemble, postprocessing, TTA/TTT, verifier, or agent planning. The codebook used during annotation defines inclusion, exclusion, evidence, status, and confidence rules for every label.

\section{Outcome Coding Examples}

The outcome is derived from the independently coded task, backbone, and output interface fields. A system whose backbone includes a decoder LLM and whose output set is exactly \{\texttt{free\_generation}\} receives outcome 0. A system with \{\texttt{classification\_head}\} or \{\texttt{ranking\_score}\} receives outcome 1. A system with \{\texttt{free\_generation}, \texttt{ranking\_score}\} is hybrid and also receives outcome 1. A system whose backbone includes a decoder LLM and whose output interface is unknown is excluded from the analysis denominator. These examples apply to the complete documented system, including intermediate interfaces. A decoder that generates a category string later parsed into a submitted label uses free generation. A dedicated head emitting scores for a fixed set of classes is coded as a classification head. If generated candidates are ranked before selection, generation is intermediate and the selected candidate is the final output. A decoder can also generate the final answer after an intermediate scoring stage; that system is also hybrid. Component attribution follows the documented implementation.

\section{Worked Annotation Examples}

Table~\ref{tab:appendix_evidence}A uses short aliases that map directly to \texttt{system\_id} in the released Kaggle annotations. The examples illustrate why task, backbone, output interface, and system method are coded independently.

Eedi-1 corresponds to \texttt{kaggle-system-0120}. Its source report documents intermediate answer generation for retrieval, decoder log probability scoring over fixed candidate options, and a final ranked list of misconception IDs. This example separates the final submission format from the interfaces used within the pipeline.

\section{Weighting and Reliability Details}

Each of the 28 eligible Kaggle competitions contributes 1--21 systems with decoder LLMs and known output interfaces. The pooled estimate is 0.723, versus 0.615 under equal competition weighting. The data package provides the registry, system annotations, and aggregate statistics; the annotation records retain the competition IDs and task labels used for aggregation.

Status agreement uses the three field states, and binary Cohen's $\kappa$ is computed for each observed label. Table~\ref{tab:agreement} reports macro and minimum label $\kappa$ together with exact set agreement. All values precede adjudication; the analysis uses consensus labels linked to source evidence.

\end{document}